\documentclass[a4paper,12pt]{article}

\usepackage[OT1]{fontenc} 
\usepackage{graphicx}
\usepackage{lfa10} %% paquet \`a r\'ecup\'erer sur le site de la conf
\usepackage{times}
\usepackage{amsmath,bm}
\usepackage{algorithm}
\usepackage{algorithmic}

\usepackage{xcolor}

\begin{document}

\newcommand\cs{\emph{crowdsourcing}}
\newcommand\Cs{\emph{Crowdsourcing}}
\renewcommand{\figurename}{Figure~}
\renewcommand{\tablename}{Tableau~}

%pour avoir les commandes en fran\c{c}ais
\renewcommand{\algorithmicensure}{\textbf{Post-conditions :}}
\renewcommand{\algorithmicend}{\textbf{fin}}
\renewcommand{\algorithmicrequire}{\textbf{Pr\'e-conditions :}}
\renewcommand{\algorithmicif}{\textbf{si}}
\renewcommand{\algorithmicthen}{\textbf{alors}}
\renewcommand{\algorithmicelse}{\textbf{sinon}}
\renewcommand{\algorithmicelsif}{\algorithmicelse\ \algorithmicif}
\renewcommand{\algorithmicendif}{\algorithmicend\ \algorithmicif}
\renewcommand{\algorithmicfor}{\textbf{pour}}
\renewcommand{\algorithmicforall}{\textbf{pour tout}}
\renewcommand{\algorithmicdo}{\textbf{faire}}
\renewcommand{\algorithmicendfor}{\algorithmicend\ \algorithmicfor}
\renewcommand{\algorithmicwhile}{\textbf{tant que}}
\renewcommand{\algorithmicendwhile}{\algorithmicend\
\algorithmicwhile}
\renewcommand{\algorithmicloop}{\textbf{boucler}}
\renewcommand{\algorithmicendloop}{\algorithmicend\
\algorithmicloop}
\renewcommand{\algorithmicrepeat}{\textbf{r\'ep\'eter}}
\renewcommand{\algorithmicuntil}{\textbf{jusqu'\`a}}

\date{}

\title{ Mod\'elisation du profil des contributeurs dans les plateformes de \cs \\[1.3em]
       Contributors profile modelization in crowdsourcing platforms
}

\author{\small
        \begin{tabular}[t]{c@{\extracolsep{4em}}c@{\extracolsep{5em}}c@{\extracolsep{6em}}c}
%%      pour trois auteurs      
%%      \begin{tabular}[t]{c@{\extracolsep{5em}}c@{\extracolsep{6em}}c}
%%      pour quatre auteurs ...
        Constance Thierry  & Jean-Christophe Dubois & Yolande le Gall & Arnaud Martin\\
        \end{tabular}
      ~\\
                \small{ Univ Rennes, CNRS, IRISA}  \\
        %${}^2$        \small{Mon Institut}  
      ~\small{Universit\'e de Rennes 1, Lannion France}\\
        \small{prenom.nom@irisa.fr}
}

\parskip 3mm
\maketitle

\thispagestyle{empty}

\doubleresume{
Le \cs~consiste \`a l'externalisation de t\^aches \`a une foule de contributeurs r\'emun\'er\'es pour les effectuer. 
La foule, g\'en\'eralement tr\`es diversifi\'ee, peut inclure des contributeurs non-qualifi\'es pour la t\^ache et/ou non-s\'erieux. 
Nous pr\'esentons ici une nouvelle m\'ethode de mod\'elisation de l'expertise du contributeur dans les plateformes de \cs~se fondant sur la th\'eorie des fonctions de croyance afin d'identifier les contributeurs s\'erieux et qualifi\'es.
}
{Expertise, \cs, th\'eorie des fonctions de croyance.}
{The crowdsourcing consists in the externalisation of tasks to a crowd of people remunerated to execute this ones.
The crowd, usually diversified, can include users without qualification and/or motivation for the tasks. In this paper we will introduce a new method of user expertise modelization in the crowdsourcing platforms based on the theory of belief functions in order to identify serious and qualificated users.
}
{Expertise, \cs, theory of belief functions}

\section{Introduction}

Le \cs~repose sur l'externalisation de t\^aches non-r\'ealisables par ordinateur \`a une foule. 
La diversit\'e des t\^aches conduit \`a distinguer diff\'erents types de \cs~d\'ecrits par Burger-Helmchen et P\'enin~\cite{burger11}. 
Nous nous int\'eressons aux plateformes d'activit\'es routini\`eres o\`u les internautes, appel\'es contributeurs, pr\'esents en grand nombre et avec des profils diversifi\'es, sont r\'emun\'er\'es pour r\'ealiser des micro-t\^aches et tout particuli\`erement r\'epondre \`a des questionnaires. 

Sur ces plateformes, il existe des contributeurs non-consciencieux, int\'eress\'es uniquement par la r\'emun\'eration et d'autres non qualifi\'es pour r\'ealiser certaines t\^aches. 
Il appara\^it comme essentiel de caract\'eriser la foule afin de traiter les r\'eponses de fa\c{c}on optimale, c'est pourquoi nous nous int\'eressons ici \`a la mod\'elisation de l'expertise d'un contributeur afin de d\'eterminer s'il s'agit ou non d'une personne consciencieuse et qualifi\'ee consid\'er\'ee comme un ''Expert''. 
Par opposition, nous consid\'erons les contributeurs non-consciencieux ou non-qualifi\'es sous la d\'enomination ''Non-Expert''.

Des premi\`eres \'etudes, d\'ecrites dans la section~\ref{sec:detExpContrib} de cet article, se sont d\'ej\`a appuy\'ees sur la th\'eorie des fonctions de croyance~\cite{Koulougli16,ouni17,rjab16} pour d\'eterminer le niveau d'expertise d'un contributeur.
Nous proposons ici une nouvelle mod\'elisation originale du type de contributeur par cette m\^eme th\'eorie des fonctions de croyance (BF), explicit\'ee dans la section~\ref{sec:modelisation}.
Afin de valider notre mod\`ele nous r\'ealisons une exp\'erimentation sur les donn\'ees d'une campagne de \cs~portant sur l'\'evaluation de la qualit\'e sonore de s\'equences audio pour lesquelles un contributeur peut indiquer la confiance, ou certitude, apport\'ee dans sa r\'eponse.
Les r\'esultats de cette exp\'erimentation sont pr\'esent\'es et analys\'es dans la section~\ref{sec:exp}.
Enfin, la section~\ref{sec:ccl} conclut cet article. 

\section{\'Etat de l'art}
\label{sec:detExpContrib}

Dans cette section nous rappelons dans un premier temps la th\'eorie des BF~\cite{dempster67}. 
Puis nous pr\'esentons son int\'er\^et ainsi que des \'etudes l'utilisant pour la d\'etermination d'experts dans le cadre du \cs. 

\subsection{La th\'eorie des fonctions de croyance}

La th\'eorie des BF permet une mod\'elisation de l'impr\'ecision et de l'incertitude de donn\'ees issues de plusieurs sources d'information. 
Dans le cas du \cs~ces diff\'erentes sources d'information sont les contributeurs $c$.
Si nous consid\'erons un probl\`eme donn\'e, l'ensemble des $n$ solutions possibles \`a ce probl\`eme est appel\'e cadre de discernement~: $\Omega = \{ \omega_1 ,... , \omega_n \} $.
Les fonctions de masse $m_c^\Omega$ utilis\'ees dans la th\'eorie des BF sont d\'efinies de $2^{\Omega}$ dans $[0,1]$ avec la condition de normalisation suivante~: % l'\'equation (1)
\begin{equation}
\label{eqnorm}
 \sum_{X \in 2^{\Omega}} m_c^\Omega(X) = 1
\end{equation}
Soit $X \in 2^{\Omega}$, la masse $m_c^\Omega(X)$ caract\'erise la croyance accord\'ee par la source $c$ \`a l'\'el\'ement $X$.
Lorsque $m_c^\Omega(X) > 0$, $X$ est appel\'e \'el\'ement focal.
Il existe deux \'el\'ements de  $2^{\Omega}$ ayant une signification particuli\`ere~: $\emptyset$ qui repr\'esente l'ouverture au monde hors du cadre de discernement et $\Omega$, l'ignorance.
Ainsi $m_i^\Omega(\emptyset)=0$ signifie que l'on consid\`ere un monde clos, les r\'eponses attendues sont donc constitu\'ees d'\'el\'ements de $\Omega$. 

%\textcolor{blue}{
Un cas particulier des BF est la fonction de masse \`a support simple (not\'ee $X^\Omega$) %, permettant d'affecter toute la masse \`a un sous-ensemble non vide de $2^{\Omega}$ 
d\'efinie par~: % l'\'equation (2). 
%}
\begin{eqnarray} 
  \left \{
  \begin{array}{l}
    m_c^\Omega(X) = 1-\omega  \mbox{ avec }  X \in 2^{\Omega} \setminus \Omega\\
    m_c^\Omega(\Omega) = \omega 
  \end{array}
  \right.
\end{eqnarray}
Pour r\'ealiser la fusion d'informations issues de $S$ sources, diff\'erents op\'erateurs de combinaisons peuvent \^etre utilis\'es, le plus couramment employ\'e \'etant l'op\'erateur de combinaison conjonctive que l'on peut \'ecrire~:  % l'\'equation (3). 
\begin{eqnarray} 
  \begin{array}{l}
    \mbox{Soit~} X,Y_1, ..., Y_S\in 2^{\Omega}\\
    \displaystyle m_{Conj}^\Omega(X)= \sum_{ Y_1 \cap...\cap Y_S  = X} \prod_{c=1}^N m_c^\Omega(Y_i)
    \label{eq:mConj}
  \end{array}
\end{eqnarray}
Celui-ci permet de diminuer l'impr\'ecision sur les \'el\'ements focaux et d'augmenter la croyance sur les \'el\'ements concordants entre les diff\'erentes sources d'information~$c$.
De plus, dans le contexte d'un monde clos, il peut \^etre pertinent d'utiliser l'op\'erateur de combinaison conjonctive de Yager~\cite{yager87} d\'efini  comme suit~:  % \'equation (4)
\begin{eqnarray} 
  \begin{array}{l}
    \mbox{Soit~} X \in 2^{\Omega}\\
    \displaystyle m_{Y}^\Omega(X)= m_{Conj}^\Omega(X), X \neq \emptyset, X \neq \Omega \\
    m_{Y}^\Omega(\Omega) = m_{Conj}^\Omega(\Omega) + m_{Conj}^\Omega(\emptyset) \\
    m_{Y}^\Omega(\emptyset) = 0
    \label{eq:mYconj}
  \end{array}
\end{eqnarray}
Lorsque les informations sont issues de plusieurs cadres de discernement $\Omega$ et $\Theta$ que l'on souhaite combiner, on r\'ealise l'extension vide sur ces cadres avant la combinaison.
Elle est donn\'ee pour tout $A\subset \Omega$ par~: % \'equation (5)
\begin{eqnarray} 
    m^{\Omega \uparrow \Omega \times \Theta}(B) =
    \left \{
    \begin{array}{l}
        m^\Omega(A) \mbox{ si } B = A \times \Theta \\
        0 \mbox{ sinon }
    \end{array}
    \right.
    \label{extension}
\end{eqnarray}
De m\^eme, consid\'erant le produit cart\'esien $\Omega \times \Theta$ on peut se projeter sur le cadre de discernement $\Theta$ (respectivement $\Omega$) en r\'ealisant une marginalisation du produit conjonctif (\'equation~\eqref{marginalisation}).
Ainsi, une fonction de masse pour la marginalisation de $\Omega \times \Theta \downarrow \Theta$ est donn\'ee 
$\forall B \subseteq \Omega$~:
%pour $X = A \times B, A \subset \Omega$ par~: 
% \'equation (6)
\begin{equation}
    \displaystyle m^{\Omega \times \Theta \downarrow \Omega}(B) = 
        \sum_{A \subseteq \Omega \times \Theta, A \downarrow \Omega = B} m^{\Omega \times \Theta}(A)
    %m^{\Omega \times \Theta \downarrow \Theta}(X) = m^{\Theta}(B).
    \label{marginalisation}
\end{equation}
Afin de prendre une d\'ecision sur les \'el\'ements du cadre de discernement $\Omega$ nous consid\'erons l'\'el\'ement focal $\omega_i \in \Omega$ pour lequel on obtient la probabilit\'e pignistique maximale $betP$~:
$$betP(\omega_i) = \max_{\omega \in \Omega}betP(\omega)$$
Avec $betP$ d\'efinie par l'\'equation~: % \'equation (7)
\begin{equation}
 \displaystyle betP(X) = \sum_{Y \in 2^{\Omega}, Y \neq \emptyset} \frac{|X \cap Y|}{|Y|} \frac{m^\Omega(Y)}{1 - m^\Omega(\emptyset)}
 \label{eq : betP}
\end{equation}
L'utilisation de cette th\'eorie dans le cadre du \cs~est int\'eressante car elle permet de mod\'eliser l'incertitude et l'impr\'ecision des r\'eponses d'un contributeur ainsi que la fusion des donn\'ees issues de l'ensemble des contributeurs vus comme des sources d'information.

\subsection{Mod\'elisations existantes}

Usuellement la m\'ethode utilis\'ee pour d\'eterminer la bonne r\'eponse \`a une question dans les plateformes de \cs~est la m\'ethode par vote majoritaire (MV) consistant \`a consid\'erer que la r\'eponse donn\'ee par la majorit\'e des contributeurs est la bonne.
Un expert, consid\'erant cette m\'ethode, est un contributeur ayant r\'epondu ''juste'' \`a un maximum de questions.
Cette m\'ethode est limit\'ee car les r\'eponses des contributeurs ont toutes le m\^eme poids et leur incertitude n'est pas consid\'er\'ee.
%Le MV, bien que tr\`es simple \`a implanter, reste limit\'e car il ne prend pas en consid\'eration l'incertitude du contributeur. 
%De plus, toutes les r\'eponses ont le m\^eme poids, que le contributeur soit qualifi\'e ou non, ce qui peut introduire un biais important dans les r\'esultats.

Une mod\'elisation plus \'evolu\'ee pour la d\'etermination d'experts dans les plateformes de \cs~repose sur l'utilisation de l'algorithme Expectation-Maximisation (EM) conjointement aux probabilit\'es.
Raykar et \emph{al.}~\cite{raykar10} utilisent ainsi une approche probabiliste se fondant sur les travaux de Dawid et Skene~\cite{dawid79} et d\'eveloppent dans leur article l'apport de cette mod\'elisation compar\'ee au MV.
Les approches probabilistes apparaissent plus int\'eressantes car elles offrent la possibilit\'e de mesurer l'incertitude sur les r\'eponses.

Dans leur \'etude portant sur la combinaison des r\'eponses sur les plateformes de \cs, Koulougli et \emph{al.}~\cite{Koulougli16} comparent trois m\'ethodes fond\'ees sur le MV, l'EM et les BF. 
Cette \'etude fait appara\^itre que les BF donnent de meilleurs r\'esultats, devant respectivement l'approche utilisant l'EM et le MV. 

Ouni {\em et al.} utilisent les BF pour mesurer le degr\'e d'expertise des contributeurs.
%Leur approche est fond\'ee sur des donn\'ees d'or collect\'ees \`a partir de questions dont la r\'eponse est connue et qui sont introduites sp\'ecifiquement dans le questionnaire. 
Leur approche est fond\'ee sur l'utilisation d'un corpus de r\'ef\'erence, \`a l'aide duquel un graphe de r\'ef\'erence orient\'e est r\'ealis\'e.
%Pour d\'efinir cette mesure, les auteurs construisent \`a l'aide du corpus, un graphe de r\'ef\'erence orient\'e. 
Puis, un graphe des r\'eponses apport\'ees aux m\^emes questions de corpus est construit pour chaque contributeur. 
Le graphe de r\'ef\'erence est compar\'e aux graphes des contributeurs en vue d'estimer leur degr\'e d'expertise. 
Cette m\'ethode offre de bons r\'esultats mais son utilisation reste contrainte \`a la pr\'esence de v\'erit\'es terrain. 

%\textcolor{blue}{
Des v\'erit\'es ̀''terrain'' sont \'egalement utilis\'ees, dans une moindre mesure, conjointement aux BF dans la mod\'elisation propos\'ee par Abassi et Boukhris~\cite{abassi18}.
La pr\'ecision des contributeurs y est estim\'ee par trois mesures, l'une utilisant des v\'erit\'es ''terrain'', la seconde le MV et la derni\`ere une log distance. 
Un {\it clustering} est r\'ealis\'e en exploitant ces mesures afin de classifier les contributeurs, puis une masse est associ\'ee \`a la r\'eponse du contributeur en fonction de sa classe.
Une combinaison par r\'eponse pour l'ensemble des contributeurs et une probabilit\'e pignistique sont finalement appliqu\'ees.
%}

Dans les plateformes de \cs, les interfaces propos\'ees commun\'ement ne permettent de recueillir que des informations pr\'ecises. Il est cependant int\'eressant d'offrir au contributeur la possibilit\'e d'introduire de l'impr\'ecision dans ses r\'eponses tel que propos\'e par Rjab {\em et al.}~\cite{rjab16}. 
Leurs travaux, fond\'es sur les BF et ne n\'ecessitant pas de v\'erit\'ees terrain, portent sur l'identification d'experts via la mod\'elisation d'un degr\'e de pr\'ecision $DP_c$~\eqref{eq:IP} et d'un degr\'e d'exactitude $DE_c$~\eqref{eq:IE} sur la r\'eponse d'un contributeur $c$. 
Soit $E_c$ l'ensemble des contributeurs, $E_{c_Q}$ l'ensemble des questions auxquelles un contributeur $c$ a r\'epondu et $\Omega_q$ le cadre de discernement associ\'e \`a la question $q$~: % l'\'equation (8)
\begin{eqnarray}
 \left \{
 \begin{array}{l}
  \displaystyle DE_c = 1 - \frac{1}{|E_{c_Q}|} \sum_{q \in E_{c_Q}} d_J(m_c^{\Omega_q} , m_{E_c|c}^{\Omega_q}  ) \\
  \displaystyle m_{E_c|c}^{\Omega_q}(X) = \frac{1}{|E_c|-1} \sum_{j \in E_c|c} m_j^{\Omega_q}(X)
 \end{array}
 \right.
 \label{eq:IE}
\end{eqnarray}
% l'\'equation (9)
\begin{eqnarray}
  \left \{
  \begin{array}{l}
  \displaystyle DP_c =  \frac{1}{|E_{c_Q}|} \sum_{q \in E_{c_Q}} \delta_c^{\Omega_q}\\
  \displaystyle \delta_c^{\Omega_q} = 1 - \sum_{X \in 2^{\Omega_q}} m_c^{\Omega_q}(X) \frac{log_2(|X|)}{log_2(|\Omega_q|)}
 \end{array}
 \right.
 \label{eq:IP}
\end{eqnarray}
Dans l'\'equation~\eqref{eq:IE} $d_J$ est la distance de Jousselme~\cite{jousselme01} entre la masse $m_c^{\Omega_q}$ et la moyenne des masses $m_{E_c|c}^{\Omega_q}$.
Un degr\'e global d'expertise $GD_{c}$ est ensuite calcul\'e en pond\'erant les degr\'es $DE_c$ et $DP_c$ par un coefficient $\beta_{c} \in [0,1]$~:  % l'\'equation (10)
\begin{equation}
 DG_{c} = \beta_{c} DE_{c} + (1-\beta_{c}) DP_{c} 
  \label{eq:rjab}
\end{equation}
Leur \'etude r\'ealis\'ee sur des donn\'ees g\'en\'er\'ees propose une comparaison avec une approche probabiliste montrant ainsi le gain d'efficacit\'e de leur m\'ethode.

\section{Caract\'erisation du contributeur}
\label{sec:modelisation}

Nous proposons ici une approche innovante utilisant certains concepts \'etablis par Rjab {\em et al.}~\cite{rjab16} afin de caract\'eriser le profil des contributeurs sans utiliser de v\'erit\'e terrain.  Cette m\'ethode pr\'esente l'int\'er\^et de pouvoir \^etre appliqu\'ee \`a l'ensemble des questionnaires propos\'es sur les plateformes de \cs.
Dans le cadre de notre mod\'elisation, le contributeur est invit\'e \`a sp\'ecifier pour chaque question pos\'ee la certitude qu'il a en la r\'eponse donn\'ee et ce afin de pouvoir \'etablir une mesure de sa confiance.
En vue de d\'efinir l'expertise du contributeur, nous nous int\'eressons non seulement \`a sa Connaissance, mais \'egalement \`a sa motivation en \'etudiant son Comportement. La Connaissance du contributeur est d\'efinie par sa Qualification reposant sur sa Confiance et l'exactitude de ses r\'eponses.

Pour \'etudier le comportement, nous nous sommes int\'eress\'es \`a la ''conscience'' d'un contributeur qui se traduit par son implication dans la r\'ealisation de sa t\^ache. 
La conscience est l'une des cinq caract\'eristiques d\'efinies dans le mod\`ele des \textit{Big Five}, aussi appel\'e mod\`ele OCEAN, propos\'e par Goldberg~\cite{goldberg93} pour caract\'eriser la personnalit\'e d'un individu. 
Les cinq \linebreak caract\'eristiques sont l'Ouverture \`a l'exp\'erience, la Conscience, l'Extraversion, l'Agr\'eabilit\'e et le N\'evrosisme.
Dans une pr\'ec\'edente \'etude, Kazai et al.~\cite{kazai11}, qui ont introduit ce mod\`ele dans le contexte du \cs~pour d\'eterminer la relation entre les traits de personnalit\'e et la qualit\'e des r\'eponses, ont conclu que la Conscience a un impact fort sur l'exactitude des r\'esultats du contributeur.
Dans notre approche, nous estimons la Conscience du contributeur en consid\'erant sa R\'eflexion et utilisons \`a cet effet le temps de r\'eponse du contributeur \`a une question.
Enfin, \`a partir de l'estimation de la Connaissance et du Comportement, nous proposons une r\'epartition des contributeurs selon deux profils~: ''Expert'' et ''Non-Expert''. 

\subsection{Connaissance du contributeur}

\subsubsection{Confiance}

Les questionnaires propos\'es dans le cadre du \cs~int\`egrent g\'en\'eralement des questions \`a choix multiples. 
Pour une question $q$, nous consid\'erons donc les r\'eponses possibles $\omega_1, \ldots, \omega_n$.
Nous d\'efinissons alors le cadre de discernement sur ces r\'eponses comme suit~: 
$\Omega_{1} = \{\omega_1,\ldots,\omega_n\}$.
Soit la r\'eponse $X \in 2^{\Omega_{1}}$ d'un contributeur $c$ \`a la question $q$, la fonction de masse \`a support simple $X^{\alpha_{c_q}}$ associ\'ee est $m_{c_q}^{\Omega_1}$, avec $\alpha_{c_q}$ la valeur num\'erique de l'incertitude du contributeur sur sa r\'eponse. 
 
\subsubsection{Qualification}

Nous d\'efinissons la Qualification comme une appr\'eciation de la valeur professionnelle d'un contributeur en fonction de l'exactitude des r\'eponses qu'il fournit aux questions pos\'ees. 
Nous consid\'erons ainsi le fait qu'un individu soit qualifi\'e ''$Q$'' ou non ''$NQ$''  pour une t\^ache avec le cadre de discernement~: $\Omega_2=\{Q, NQ\}$. 
Comme nous nous reposons sur l'exactitude des r\'eponses du contributeur pour estimer sa Qualification, nous associons \`a la masse $m_c^{\Omega_2}$ le degr\'e d'exactitude $DE_c$. 
Aussi pour l'\'equation~\eqref{eq:IE} nous avons $m_c^{\Omega_q} = m_{c_q}^{\Omega_1}$. 
Avec une masse nulle pour l'ignorance, nous associons  $DE_c$ \`a l'\'el\'ement focal ''$Q$'' et $1-DE_c$ \`a ''$NQ$''. Afin de ne pas \^etre cat\'egorique sur la Qualification du contributeur, nous interpr\'etons l'impr\'ecision sur ses r\'eponses aux questions comme de l'ignorance.
Nous affaiblissons ainsi l'information donn\'ee par le contributeur avec le degr\'e de pr\'ecision $DP_c$ donn\'e par l'\'equation~\eqref{eq:IP} o\`u $\Omega_q = \Omega_1$. 
Nous obtenons alors~: % l'\'equation (11)
\begin{eqnarray} 
    \left \{
    \begin{array}{l}
        \displaystyle m_{c}^{\Omega_2}(Q)= DP_c * DE_c  \\
        \displaystyle m_{c}^{\Omega_2}(NQ) = DP_c * (1 - DE_c)\\
        m_{c}^{\Omega_2}(\Omega_2) =1 - DP_c
    \end{array}
    \right.
\end{eqnarray}
 
\subsection{Comportement du contributeur}

\subsubsection{R\'eflexion}
Nous nous int\'eressons ici au temps de r\'eflexion pris par le contributeur pour donner sa r\'eponse.
Consid\'erons le cadre de discernement suivant~: 
$\Omega_3 = \{R, NR\}$, 
o\`u ''$R$'' signifie que la r\'eponse du contributeur est r\'efl\'echie et ''$NR$'' qu'elle est instinctive.
Soit un \'el\'ement $X \in 2^{\Omega_3}$ indiquant la R\'eflexion du contributeur $c$ pour une question $q$, nous d\'efinissons la masse associ\'ee \`a $X$ par~: % l'\'equation (12)
\begin{equation}
 m_{c_q}^{\Omega_3}(X) = g(T_{c_q},T_{th_q},X)
 \label{m_4}
\end{equation}
%\textcolor{blue}{
La fonction $g$ est d\'efinie par le pseudo-code~\ref{algo:g}, $T_{c_q}$ est le temps de r\'eponse du contributeur $c$ \`a la question $q$ et $T_{th_q}$ un temps de r\'eponse th\'eorique attendue \`a $q$. 
La masse $m$ est initialis\'ee par une constante, la r\'eflexion \`a instinctive (NR).
La fonction $alpha$ retourne une valeur entre 0 et 1 proportionnelle aux temps renseign\'es en param\`etres.
Consid\'erons le fait que nous pouvons avoir un nombre important de r\'eponses \`a combiner nous n'utilisons pas l'op\'erateur de Yager, car pour de nombreuses sources il pourrait engendrer davantage de conflits et il en r\'esulterait une masse trop importante sur l'ignorance.
Aussi, nous consid\'erons la moyenne des masses~\eqref{m_4} sur $q$~: $m_{c}^{\Omega_3}$.
%}
\begin{algorithm} [t]
  \caption{
    Fonction $g$(r\'eel:$T_{c_q},T_{th_q}$, 
    \newline caract\`ere: $X$)
   }
  \label{algo:g}
\begin{algorithmic}

\STATE caract\`ere $reflexion \leftarrow NR$
\STATE r\'eel $m \leftarrow C^{te}$
\STATE r\'eel $\alpha_3 \leftarrow alpha(T_{c_q},T_{th_q})$
\IF{($T_{c_q} > T_{th_q} $)}
\STATE $reflexion \leftarrow R $ 
\ENDIF

\IF{($X = reflexion$)}
\STATE $m \leftarrow \alpha_3 $ 
\ELSIF{($X = \Omega_3$)}
\STATE $m \leftarrow 1 - m - \alpha_3 $ 
\ENDIF

\RETURN $m$
\end{algorithmic}
\end{algorithm}

\subsection{Profil du contributeur}  

Nous pouvons finalement aboutir dans cette section \`a la mod\'elisation de l'expertise d'un contributeur.
Pour ce faire, nous consid\'erons qu'un ''Expert'' est un contributeur qualifi\'e pour la t\^ache et r\'efl\'echit dans la r\'ealisation de celle-ci. 
%Nous restreignons ici la notion de Conscience \`a la R\'eflexion du contributeur lors de la r\'ealisation de la t\^ache.
Nous d\'efinissons le cadre de discernement de l'expertise d'un contributeur comme le produit des cadres de discernement de la Qualification et de la R\'eflexion~: $\Omega_4 = \Omega_2 \times \Omega_3$. 
Afin d'\'etablir la masse $m_c^{\Omega_2 \times \Omega_3}$ nous commen\c{c}ons par d\'efinir les masses $m_c^{\Omega_2 \uparrow \Omega_2 \times \Omega_3}$ et $m_c^{\Omega_3 \uparrow \Omega_2 \times \Omega_3}$ de mani\`ere analogue \`a l'\'equation~\eqref{extension}. 
Nous appliquons ensuite l'op\'erateur conjonctif de Yager~\eqref{eq:mYconj} \`a ces deux masses afin d'obtenir $m_c^{\Omega_4}$. 

\section{Exp\'erimentations}
\label{sec:exp}

Dans le cadre de cette \'etude une campagne de \cs~a \'et\'e r\'ealis\'ee sur l'\'ecoute d'enregistrements sonores pour lesquels les contributeurs devaient donner une note entre 1 et 5. 
Cette campagne \'etait constitu\'ee de 4 HITs (Human Intelligence Tasks) comprenant 12 questions chacun parmi lesquelles figuraient 5 MNRUs (Modulated Noise Reference Units). 
Les MNRUs sont des enregistrements dont la qualit\'e sonore est modul\'ee par du bruit. 
Il s'agit ici de v\'erit\'ees terrain dont on conna\^it la qualit\'e relative, \'echelonn\'ee de 1 \`a 5. 
Le premier enregistrement est ainsi de qualit\'e mauvaise et le cinqui\`eme de qualit\'e excellente. 
La diff\'erence de qualit\'e sonore entre les MNRUs est beaucoup plus importante que celle qui existe entre les 7 autres enregistrements, nomm\'ees par la suite donn\'ees de test, dont la qualit\'e \'etait non connue et plus difficile \`a \'evaluer. Une foule de 93 contributeurs devait \'ecouter chacun des enregistrements dans un ordre al\'eatoire et sp\'ecifier leur qualit\'e. Leurs r\'eponses pouvaient \^etre impr\'ecises, avec la possibilit\'e de donner deux niveaux de qualit\'e cons\'ecutifs, et \'etaient associ\'ees \`a un degr\'e de certitude indiquant la confiance sur le r\'esultat fourni. 

\subsection{Approche propos\'ee}  
Nous ignorons la Qualification et la Conscience des contributeurs sollicit\'es en \cs~, nos travaux visent donc \`a d\'eterminer leur expertise r\'eelle. 
Dans cet objectif, nous commen\c{c}ons par calculer la masse $m_c^{\Omega_2}$ en \'evaluant tout d'abord $m_{c_q}^{\Omega_1}$. 
Les valeurs $\alpha_{c_q}$ utilis\'ees sont sp\'ecifi\'ees dans le tableau~\ref{tab:alpha_u_q}.  
\begin{table}[t]
  \centering
  \footnotesize{
  \begin{tabular}{|l|l|}
	\hline
     R\'eponse & $\alpha_{u_q}$ \\
        \hline
        Tr\`es S\^ur~:& $\alpha_{N5} = 0.99 $ \\
        Plut\^ot S\^ur~:& $\alpha_{N4} = 0.75$ \\ 
        Moyennement S\^ur~:& $\alpha_{N3} = 0.5$ \\
        Peu S\^ur~:& $\alpha_{N2} = 0.25$ \\
        Pas S\^ur~:& $\alpha_{N1} = 0.01$ \\
        \hline
    \end{tabular}
    }
    \caption{Incertitudes et valeurs associ\'ees}
    \label{tab:alpha_u_q}
\end{table}
Pour ces valeurs de $\alpha_{c_q}$ nous avons choisi de ne pas associer une valeur $\alpha_{N5}=1$ \`a une r\'eponse indiqu\'ee comme ''Tr\`es S\^ur'' (r\'eciproquement $\alpha_{N1}\neq0$ pour une r\'eponse ''Pas S\^ur''). En effet, m\^eme si le contributeur accorde une confiance absolue dans sa r\'eponse, nous pr\'ef\'erons maintenir une incertitude en l'absence de connaissance relative \`a l'expertise du contributeur. 
Ensuite nous \'etablissons les masses $m_c^{\Omega_2}$ et $m_{c}^{\Omega_3}$ pour d\'eterminer l'expertise du contributeur repr\'esent\'ee par $m_c^{\Omega_4}$.

\subsection{Validation de l'approche}  
Nous r\'ealisons notre \'etude sans int\'egrer les v\'erit\'es terrain (MNRUs) dans le corpus d'exp\'erimentation en utilisant uniquement les donn\'ees de test. Cependant, pour valider l'approche, nous r\'ealisons par ailleurs une estimation th\'eorique de l'expertise du contributeur uniquement \`a partir des MNRUs et \'etudions la convergence entre les diff\'erents r\'esultats obtenus. 

\begin{figure}[t]
  \begin{center}
  \includegraphics[scale=0.4]{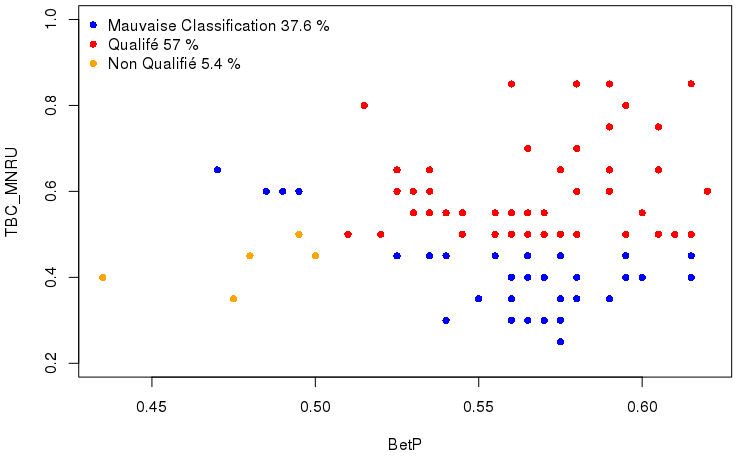} 
  \end{center}
  \caption{BetP sur les MNRUs}
  \label{fig:m3_mnru} 
\end{figure}

\begin{figure}[t]
  \begin{center}
  \includegraphics[scale=0.4]{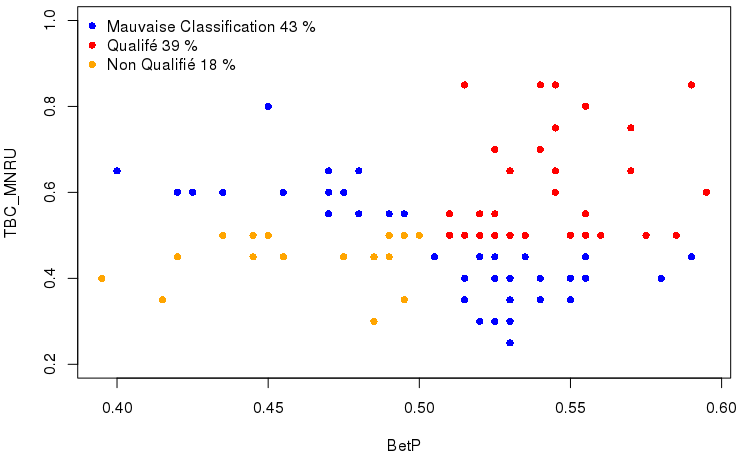} 
  \end{center}
  \caption{BetP sur les donn\'ees de test}
  \label{fig:m3_6_12} 
\end{figure}

Pour \'etablir cette estimation th\'eorique, une matrice de confusion est r\'ealis\'ee afin de mesurer les \'ecarts entre les valeurs attendues pour les MNRUs et les r\'eponses attribu\'ees par les contributeurs \`a ces s\'equences. 
Cette matrice nous permet de calculer le taux de bonne classification du contributeur sur les 5 MNRUs ($TBC_{M5}$) ce qui s'apparente \`a un taux de bonne r\'eponse pour le contributeur.
Nous appliquons une probabilit\'e pignistique~\eqref{eq : betP} \`a la masse portant sur la Qualification $m_c^{\Omega_2}$ dont nous noterons la valeur BetP, ce qui nous permet d'obtenir la figure~\ref{fig:m3_mnru} en ne consid\'erant que les r\'eponses du contributeur aux MNRUs et la figure~\ref{fig:m3_6_12} pour les r\'eponses aux donn\'ees de test.
Nous consid\'erons un seuil $\sigma = 0.5$ au-dessus duquel un contributeur est consid\'er\'e comme ''Expert''.
Ainsi si $TBC_{M5}>\sigma$ le contributeur est consid\'er\'e th\'eoriquement comme ''Expert''  et si $BetP>\sigma$ le contributeur est exp\'erimentalement consid\'er\'e comme ''Expert''.
En revanche, si $TBC_{M5}>\sigma$ et $BetP\leq\sigma$ le contributeur th\'eoriquement ''Expert'' est singuli\`erement classifi\'e comme ''Non-Expert'' par notre mod\'elisation.\\
Consid\'erons ces deux figures, nous constatons un nombre plus important de contributeurs classifi\'es comme ''Expert'' que ''Non-Expert'' par BetP. 
En l'absence de v\'erit\'ees terrain pour \'etablir l'expertise des contributeurs, les pourcentages de bonne classification de 62.4\% pour la figure~\ref{fig:m3_mnru} et de 57\% pour la figure~\ref{fig:m3_6_12} sont positif.
En effet, comme la t\^ache d\'epend de l'appr\'eciation du contributeur et que les fluctuations dans la d\'egradation sonore des donn\'ees ne sont pas \'evidentes \`a percevoir, ces \'el\'ements peuvent expliquer la diff\'erence entre les r\'esultats attendus pour les MNRUs et ceux obtenus.
De plus si nous consid\'erons les r\'esultats de classification de la figure~\ref{fig:m3_mnru} comme r\'ef\'erence et que nous comparons leur intersection avec ceux de la figure~\ref{fig:m3_6_12}, nous avons 85.85 \% de bonne classification en commun. Cela signifie que nos r\'esultats restent \`a un bon niveau de fiabilit\'e en consid\'erant des questions diff\'erentes, MNRUs ou donn\'ees de test.
\begin{figure}[t]
  \begin{center}
  \includegraphics[scale=0.4]{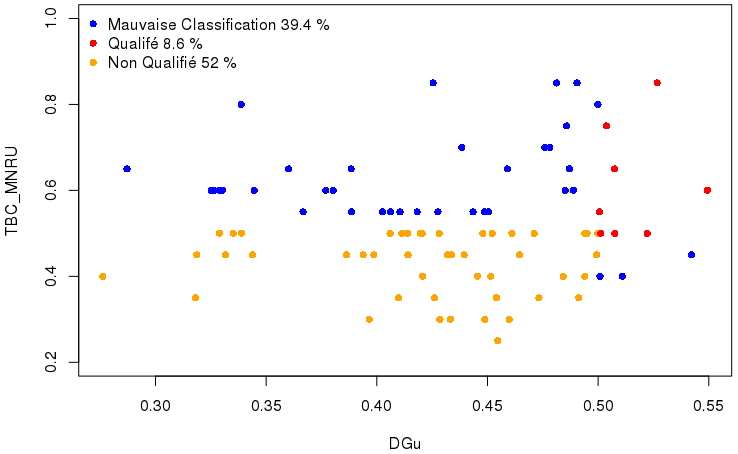} 
  \end{center}
  \caption{$DG_c$ sur les MNRUs}
  \label{fig:d_Rajb_mnru} 
\end{figure}

Notre mesure de la Qualification du contributeur ($m_c^{\Omega_2}$) reposant sur les degr\'es $DE_c$~\eqref{eq:IE} et de $DP_c$~\eqref{eq:IP} de Rjab {\em et al.}~\cite{rjab16}, nous comparons \'egalement les r\'esultats de notre mod\'elisation avec ceux obtenus par ce mod\`ele ~\eqref{eq:rjab} pour la valeur de $\beta_{c} = 0.5$. En mesurant le degr\'e global de Rjab  $DG_c$ pour l'ensemble des MNRUs nous obtenons la figure~\ref{fig:d_Rajb_mnru}. 
%\textcolor{blue}{
Nous pouvons constater avec le m\^eme seuil $\sigma = 0.5$ que la majorit\'e des contributeurs (88.2\%) est consid\'er\'ee comme ''Non-Expert'' par ce mod\`ele, puisque 11 contributeurs seulement (11.8\%) poss\`edent un $DG_u$ sup\'erieur \`a $\beta_{c}$.
%}

Il est \'egalement int\'eressant de pouvoir comparer nos diff\'erents r\'esultats, la Qualification $\Omega_2$, la R\'eflexion $\Omega_3$ et l'Expertise $\Omega_4$, avec le degr\'e global $DG_c$. 
A cet effet, nous projetons par marginalisation~\eqref{marginalisation} les r\'esultats de $\Omega_4$ sur $\Omega_2$. Les tableaux~\ref{tab:TBC_mnru} et~\ref{tab:TBC_6_12} contiennent les taux de bonne classification des contributeurs ($TBC_c$) en consid\'erant comme ''Expert'' les contributeurs pour lesquels $TBC_{MNRU} > \sigma$ pour les r\'esultats attendus (MNRUs) et $DG_c ,BetP > \sigma$ pour les donn\'ees de test.
\begin{table}[t]
\centering
 \begin{tabular}{|c|c|c|c|}
 %\hline
  %\multicolumn{4}{|c|}{MNRU}\\
  \hline
  $m_c^{\Omega_2}$ & $m_{Moy\_c}^{\Omega_3}$ & $m_c^{\Omega_4}$  &$DG_c$ \\
 \hline
 62.4 \% & 58.0 \% & 62.4 \% & 60.6 \%  \\
 \hline
 \end{tabular}
 \caption{$TBC_c$ pour les MNRUs}
 \label{tab:TBC_mnru}
\end{table}
\begin{table}[t]
\centering
 \begin{tabular}{|c|c|c|c|}
 %\hline
  %\multicolumn{4}{|c|}{6-12} \\
  \hline
  $m_c^{\Omega_2}$ & $m_{Moy\_c}^{\Omega_3}$ & $m_c^{\Omega_4}$& $DG_c$ \\
 \hline
 57.0 \% & 58.0 \% & 57 \% & 58.0 \%\\
 \hline
 \end{tabular}
 \caption{$TBC_c$ pour les donn\'ees de test} 
 \label{tab:TBC_6_12}
\end{table}\\
Le pourcentage de $TBC_c$ \`a partir de la R\'eflexion du contributeur est identique pour les MNRUs et les donn\'ees de test (tableaux~\ref{tab:TBC_mnru} et~\ref{tab:TBC_6_12}).
Nous pouvons en d\'eduire que le contributeur porte la m\^eme r\'eflexion \`a l'ensemble des questions et donc que son comportement est identique tout au long de la r\'ealisation de sa t\^ache. 

Le $TBC_c$ pour la Qualification du contributeur $m_c^{\Omega_2}$ est en revanche plus important pour les MNRUs que pour les donn\'ees de test. Cette diff\'erence est li\'ee \`a l'\'echelle de d\'egradation sonore plus importante pour les MNRUs que pour ces autres donn\'ees. 

Le m\^eme raisonnement que pour $m_c^{\Omega_2}$ s'applique \`a $m_c^{\Omega_4}$ et au degr\'e $DG_c$, car ces trois mesures reposent sur les degr\'es $DE_c$ et $DP_c$, ce qui explique la diminution de leur $TBC_c$ du tableau~\ref{tab:TBC_mnru} au tableau~\ref{tab:TBC_6_12}.
Une am\'elioration significative du $TBC_c$ sur $\Omega_4$ compar\'e \`a $\Omega_2$ aurait pu \^etre observ\'ee mais nous supposons que la R\'eflexion du contributeur \'etant constante, elle n'apporte pas suffisamment d'information pour l'estimation de son degr\'e d'expertise. 
Notre mesure de l'expertise offre un $TBC_c$ proche de celui de $DG_c$, sup\'erieur pour les MNRUs et plus faible pour les autres donn\'ees.

Malgr\'e la difficult\'e de proposer une mod\'elisation pertinente de l'expertise des contributeurs en l'absence de v\'erit\'e terrain, notre approche apporte des r\'esultats int\'eressants au vu des exp\'erimentations men\'ees. 

\section{Conclusion et perspectives}
\label{sec:ccl}
Le \cs~repose sur l'externalisation de t\^aches \`a une foule de contributeurs. 
Cette foule \'etant g\'en\'eralement diversifi\'ee, il est important de caract\'eriser au mieux les individus qui la compose en vue d'une meilleure exploitation des donn\'ees.
Ouni et al.~\cite{ouni17} ont propos\'e une mod\'elisation de l'expertise des contributeurs utilisant des v\'erit\'es terrain.
N\'eanmoins cette mod\'elisation est limit\'ee car il n'est pas toujours possible de disposer de telles donn\'ees.
La mod\'elisation que nous proposons pr\'esente l'int\'er\^et de ne pas n\'ecessiter de v\'erit\'e terrain et a donc pour objectif d'\^etre plus facilement exploitable.
Rjab et al.~\cite{rjab16} ont \'etudi\'e sur des donn\'ees g\'en\'er\'ees, une mod\'elisation faisant abstraction des v\'erit\'es terrain et mesurant un degr\'e d'expertise global du contributeur.
Ce degr\'e repose sur la mesure de degr\'es d'exactitude et de pr\'ecision.
Dans notre mod\'elisation, nous exploitons ces degr\'es significatifs de la Connaissance du contributeur.
Afin d'affiner notre mesure d'expertise, nous prenons \'egalement en consid\'eration son Comportement dans la r\'ealisation des t\^aches, \`a travers sa R\'eflexion.

Des exp\'erimentations ont \'et\'e r\'ealis\'ees sur des donn\'ees de test, puis une validation de nos r\'esultats a \'et\'e faite avec des v\'erit\'e terrain. Nos exp\'erimentations concordent en partie \`a nos attentes. Nous attribuons les erreurs de classification \`a une estimation incompl\`ete de la Connaissance du contributeur ainsi qu'\`a un manque d'information sur son Comportement. 
Afin d'am\'eliorer l'estimation de la Connaissance, nous consid\'erons que la Qualification seule ne suffit pas et qu'il serait int\'eressant de consid\'erer davantage la t\^ache et sa difficult\'e. Nous envisageons dans la suite de notre travail de permettre au contributeur d'exprimer, non seulement sa confiance sur ses r\'eponses, mais aussi dans quelle mesure la t\^ache lui para\^it int\'eressante ou complexe. Cela permettrait de mesurer son int\'er\^et ainsi que son aisance \`a la mener \`a bien.
Enfin, pour mod\'eliser le Comportement, nous nous sommes limit\'es \`a la R\'eflexion mais il serait int\'eressant de prendre d'autres crit\`eres en compte, tels que l'attention du contributeur, afin de parfaire notre mod\`ele.

\end{document}